\documentclass[conference]{IEEEtran}
\IEEEoverridecommandlockouts
\usepackage{cite}
\usepackage{amsmath,amssymb,amsfonts}
\usepackage{algorithmic}
\usepackage{graphicx}
\usepackage{textcomp}
\usepackage[linesnumbered,ruled]{algorithm2e}
\usepackage{array}
\usepackage{subfigure}

\newlength\mylen

\newcolumntype{M}[1]{>{\centering\arraybackslash}m{#1}}
  
\def\BibTeX{{\rm B\kern-.05em{\sc i\kern-.025em b}\kern-.08em
    T\kern-.1667em\lower.7ex\hbox{E}\kern-.125emX}}
\begin{document}

\title{Progressively Growing Generative Adversarial Networks for High Resolution Semantic Segmentation of Satellite Images \\
{}
}

\author{\IEEEauthorblockN{Edward Collier$^1$, Kate Duffy$^2$, Sangram Ganguly$^3$, Geri Madanguit$^4$, Subodh Kalia$^4$, Gayaka Shreekant$^4$, \\Ramakrishna Nemani$^5$, Andrew Michaelis$^5$, Shuang Li$^5$, Auroop Ganguly$^2$, Supratik Mukhopadhyay$^1$}
\IEEEauthorblockA{\textit{Louisiana State University$^1$, Northeastern University$^2$, NASA Ames Research Center/BAERI$^3$,}\\ \textit{BAER Institute$^4$, NASA Ames Research Center$^5$}}
}

\maketitle

\begin{abstract}Machine learning has proven to be useful in classification and segmentation of images. In this paper, we evaluate a training methodology for pixel-wise segmentation on high resolution satellite images using progressive growing of generative adversarial networks. We apply our model to segmenting building rooftops and compare these results to conventional methods for rooftop segmentation. We present our findings using the SpaceNet version 2dataset. Progressive GAN training achieved a test accuracy of 93\% compared to 89\% for traditional GAN training.

\end{abstract}

\section{Introduction}
Due to the massive, and increasing, amount of satellite data available, a significant effort has been devoted to developing machine learning methods for satellite image processing. Among the higher level products sought, rooftop detection has received particular attention due to the diverse insights available from rooftop products. Rooftop detection is used to track urban growth, estimate population, assess damage from natural disasters and classify land use, among other applications.

Training rooftop segmentation models presents challenges, like the similar appearance of rooftops to other objects such as cars. Rooftops also have dissimilar appearances from city to city. Building shape, building material and surrounding land cover vary widely from scene to scene and present challenges for transfer of models between cities. As such, no generalizable model yet exists that can accurately detect roofs in the full population of satellite images. In the application of models to diverse scenes, a tradeoff exists between accuracy and generalization.

Despite challenges, rooftop products have important applications in the Earth sciences, like in studies of the urban heat island effect. Climate projections indicate that throughout the 21st century regional heat waves will become more frequent, intense and long-lasting \cite{Meehl994}. This trend in heat waves results from shifts in circulation patterns driven by climate forcings including greenhouse gases, volcanic aerosols and solar variability. In urban zones, the heat island effect is well known to exacerbate heat waves. Under this effect, urban areas experience increased ambient temperatures when compared to nearby rural locations. Among other impacts, urban heat islands have been linked to increased summer mortality rates \cite{Tan2010}.   

Compounded by the global trend toward urbanization, urban heat islands threaten public health, ecosystems, and economies. Rooftop products can generate insights in the dynamics of urban heat islands along with of high-risk cities. In Figure 1 one study uses satellite imagery to explore the the relationship between land surface temperature and land use \cite{CHEN2006133}. Rooftop products can be used to determine where cities are growing most quickly and most densely \cite{SONG20161}. High-resolution rooftop products are needed for 3D city modeling at the scale of individual buildings.

\begin{figure}[ht]
\centering
\includegraphics[width=0.45\textwidth]{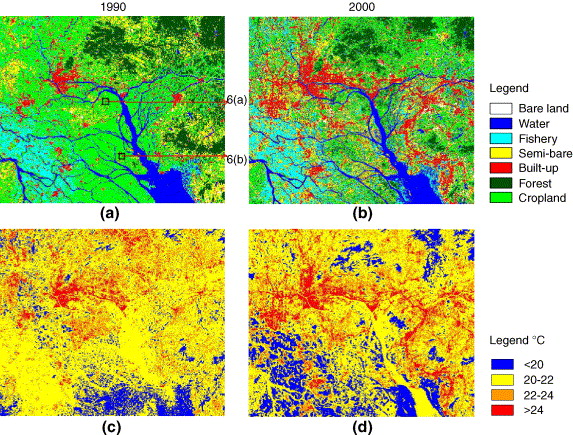}
\caption{Land use/land cover and temperature distribution patterns between 1990 and 2000. Reproduced from Chen et al., 2006 \cite{CHEN2006133}}
\end{figure}

Rooftop products are also used in post-disaster damage assessment. Remote sensing provides one of the fastest, lowest cost methods to gather information about damaged areas. Humanitarian efforts and recovery planning can be informed by location and extent of damage, identified by changes between rooftop products before and after an event \cite{DONG201385}. Automated generation of rooftop products creates a running inventory of assets, which can be leveraged to track damages.

These applications require high-resolution rooftop products. However, generation of high-resolution images presents challenges for traditionally trained deep neural networks. In this paper, we evaluate the efficacy of progressive training of a generative adversarial network (GAN) for rooftop segmentation using multi-spectral satellite images. This is, to the best of our knowledge, the first results of progressive training for semantic segmentation. The GAN consists of a generator and a discriminator, which are linked through an adversarial training algorithm. The generator learns to generate mappings from input to target and the discriminator learns to evaluate them. Feedback from the discriminator enables the generator to produce highly realistic outputs. We employ U-Net architecture, a convolutional neural network consisting of an encoder-decoder, as the generator. We apply progressive growing of the generator and discriminator. In this transfer learning process, increasingly deep networks are trained to learn increasingly complex features. Accuracy of rooftop classification is assessed and results are compared with those of a traditionally trained generative model and with those of non-generative U-Net. Our progressively trained GAN approach beats both traditional GAN and non-generative U-Net in accuracy, by four percent and eight percent respectively.
CGMNMSPSa
\begin{figure}[h!]
\centering
\includegraphics[width=0.23\textwidth]{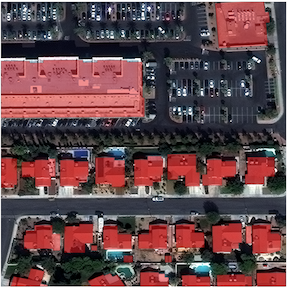}
\includegraphics[width=0.23\textwidth]{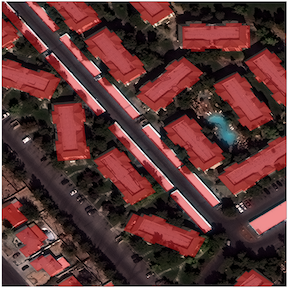}
\caption{Example labeled images from the Las Vegas Space Net dataset. Red labels reflect ground truth rooftop area.}
\end{figure}

\begin{figure*}[tb!]
\centering
\includegraphics[width=0.8\textwidth]{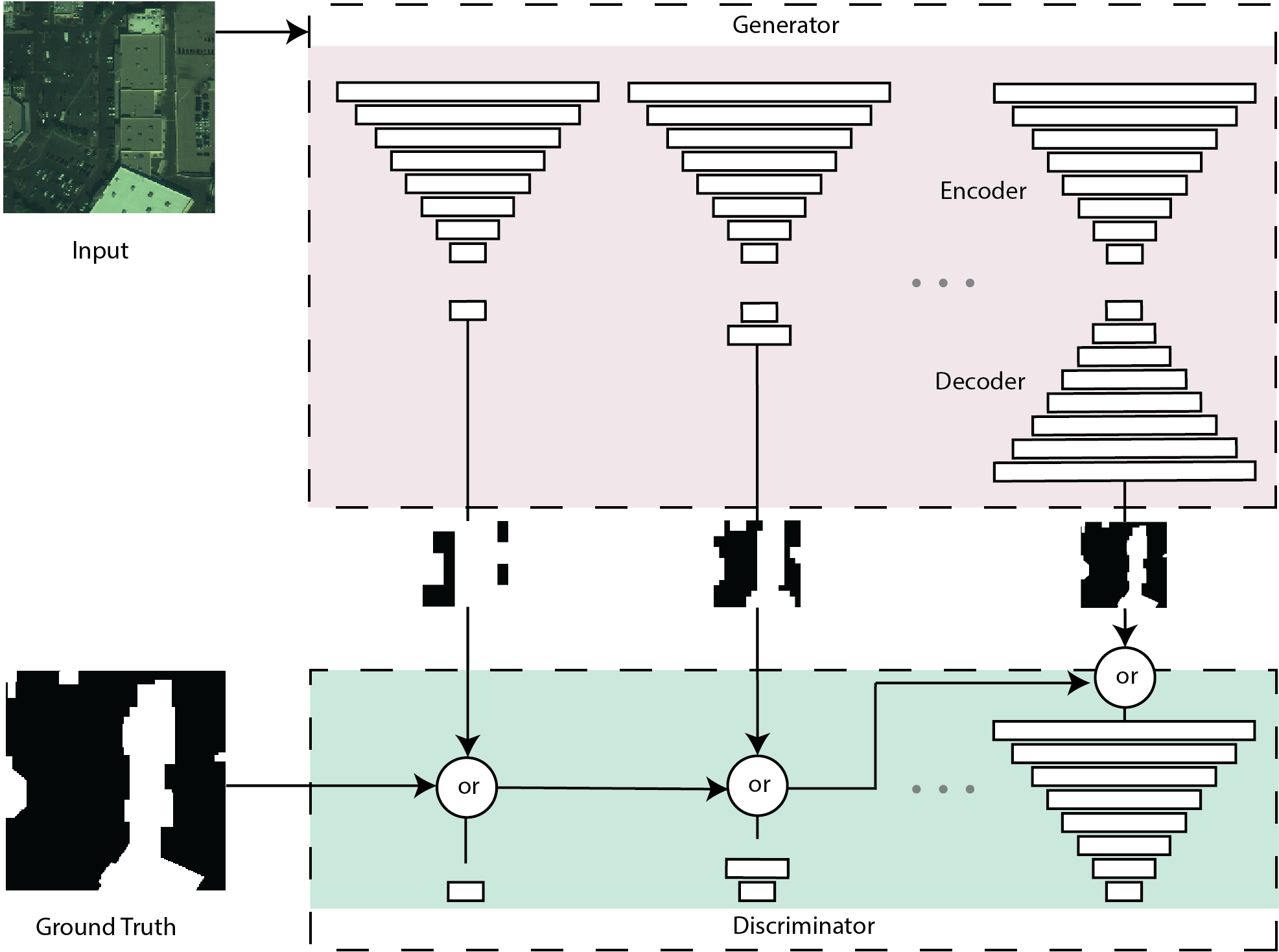}
\caption{Schematic diagram of proposed model that contains U-Net architecture as the generator. The decoder in the generator and the discriminator grow layer by layer and spatial resolution of output increases as training advances from left to right.}
\end{figure*}

\section{Related Work}\label{rw}
Significant accomplishments have been made in computer vision, resulting in increasingly effective state of the art methods for image processing \cite{DBLP:conf/icann/KarkiDBM17,DBLP:conf/ijcnn/BasuKMGNDG16}. Early efforts in automatic rooftop segmentation relied on techniques to generate candidate rooftops and subsequent evaluation to accept or reject candidate rooftops. Edge detection, corner detection, and segmentation into homogeneous regions via k-means clustering or support Vector Machines (SVM) have been used to identify candidate rooftops \cite{Baluyan2013,joshi2014roof}. Discriminative features used to evaluate candidate rooftops include building shadows, geometry and spectral characteristics \cite{ren2009, Jin2005}. Several approaches have used LIDAR alone or in addition to multi-spectral images \cite{wang2011lidar, sohn2012implicit,Bittner_2018_CVPR_Workshops}

Newer-generation machine learning techniques \cite{DBLP:journals/npl/BasuKGDMGKN17} have also been applied in satellite image classification \cite{DBLP:journals/tgrs/BasuGNMZMMVDDCY15} and in rooftop segmentation specifically \cite{basu2015deepsat, leeusing, chen2018aerial}. Convolutional neural networks (CNNs) have greatly improved the state of the art in semantic segmentation tasks wherein each pixel in an image is associated with a class label \cite{Long_2015_CVPR, chen2014}. High-resolution rooftop detection presents a dense prediction problem in which proper pixel-wise labeling is paramount to a produce a product with well-defined rooftops. Recent work in precipitation downscaling uses stacked CNNs to outperform a suite of machine learning methods \cite{DeepSD:Vandal:2017:DGH:3097983.3098004}. 
Another study uses stacked U-Nets which enhance the results of the previous U-Net \cite{khalel2018automatic}. This study found that stacking of just two CNNs outperforms the state of the art method. Introduced in 2015, U-Nets utilize skip connections and an encoder-decoder structure to learn a latent translation from input to output \cite{ronneberger2015unet}. 

CNN performance is sometimes hampered by blurry results, which satisfy the loss function by reducing the Euclidean distance between predictions and the target \cite{pathak2016context}. It is difficult to conceptualize a loss function that enforces sharpness and perceptual similarity between images. Generative adversarial networks (GAN) address this pitfall by simultaneously training a discriminator network to differentiate between real and generated images \cite{Goodfellow-GAN}. GANs hallucinate structure where it does not exist, generating sharp outputs. The GAN algorithm for image generation is further improved upon by progressively grown GANs \cite{nvidia2017progressive}. In working with high-resolution images, GANs run into issues with real and generated images being too easy to discriminate. Progressively grown GANs address this challenge by utilizing transfer learning in deep neural networks \cite{yosinski2014transferable}. Deep neural networks learn features layer by layer in a generic to specific manner. Naturally, features learned from one dataset overlap and can be applied to other data sets \cite{collier2018cactusnets}. Progressively trained GANs continually transfer the lower resolution features learned to successive steps allowing each layer to learn fine grade details individually instead of in conjunction with the rest of the layers. This progressive learning of low to high level features results in better quality images and reduces memory load for high resolution image processing.

Adversarial training has been demonstrated to improve the labeling accuracy of semantic segmentation models \cite{luc2016semantic}. 
One study compares the performance of U-Net and adversarial-trained U-Net for semantic segmentation of roads and finds that adversarial training reduces over fitting and improves validation accuracy. \cite{road-detection:unet:cgan}. Much work has been done applying adversarial semantic segmentation in medical imaging as well as style transfer \cite{kamnitsas2017brain, zhuadversarial, zhu2017unpaired}. One study focusing on rooftop segmentation uses GANs to overcome missing data, while another uses conditional GANs to refine 3D building shapes \cite{bischke2018overcoming,Bittner_2018_CVPR_Workshops}. In our study, we build upon progress in semantic segmentation by applying and evaluating progressive training.

\section{Data Preparation}
Our experiments are run on the SpaceNet version 2 dataset provided for free usage by DigitalGlobe \cite{spacenet}. This dataset contains high resolution commercial satellite images of Las Vegas, Paris, Shanghai, and Khartoum along with the masks of building and road footprints, as decpicted in Figure 2. The following experiments are run on the Las Vegas dataset for rooftop segmentation. We leave other datasets and road segmentation for future work and evaluation. 

\section{Proposed Method}\label{pm}
Our training algorithm incorporates two primary components: adversarial training and progressive growing. Our method is unique to previous works in progressive growing due to the architecture of the generator and discriminator \cite{nvidia2017progressive}. In previous works the generator and discriminator mirror one another; in our model the generator instead has an encoder-decoder structure. Our proposed model's architecture and progressive growth are presented in Figure 3.

\subsection{Network Architecture}
Many out of the box segmentation models use the U-Net architecture because of its ability to learn a latent translation between the input and target sets. Additionally, mirrored layers in U-Net contain skip connections that allow structural information to be preserved when decoding from the learned latent encoding. This architecture has become a common generator structure in many domains of GANs. It is for these reasons along with its popularity that we have chosen to use the U-Net architecture in our framework as well.

\begin{figure*}[t!]
\centering
\includegraphics[width=1\textwidth]{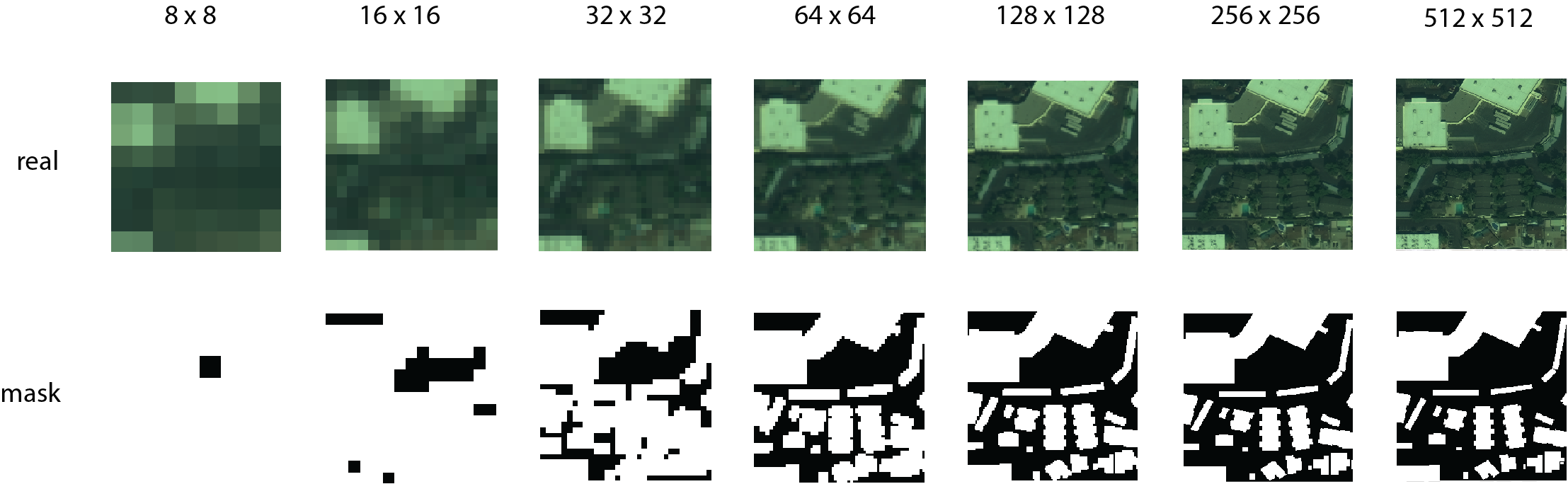}
\caption{Progress of training at different layers and resolutions}
\end{figure*}

\subsection{GAN training}
In the most basic form of a GAN, the generator learns a mapping of $z \rightarrow y$, where $z$ is some random latent vector that is translated onto the feature space defined by the task $y$. If a GAN is being used to translate one image to another, then the task of the generator is to learn a mapping $x \rightarrow y$ from input set $x$ to target $y$. This is done by mapping $x$ to latent encoding $z$, $x \rightarrow z$, which can be decoded to $y$, $z \rightarrow y$. In our case we seek to learn some mapping between a high resolution satellite image and the rooftop segment of that image.

GANs learn these mappings between inputs and targets via a min/max game, $minmaxL(G, D)$, played between the generator $G$ and the discriminator $D$ with loss $L(G, D)$. We express the GAN's objective function as:

\begin{align*}
minmaxL(G, D) = \mathbb{E}_{y}[logD(y)] +\\ \mathbb{E}_{x,z}[log(1 - D(x|G(x|z)))].
\end{align*}

Other translation objectives like conditional GANs could be used here, but those work best when the translation is from simple to complex. In our case we are translating from complex images to simple masks. In this task, non-conditional GAN loss has been demonstrated to perform best \cite{P.Isola-Cond-GAN}.

In the case of segmentation, we desire the outputs of our generator to provide the best possible. To do this we add the $L_1$ distance to the objective:

\begin{align*}
L_1(G) = \mathbb{E}_{x,y,z}[|y - G(x|z)|].
\end{align*}

This imposes a second objective for generator's output: to mirror the ground truth by forcing it to minimize the absolute distance between the two. Absolute error ($L_1$ distance) is used rather than mean squared error ($L_2$ distance) to discourage blurring.

\subsection{Progressive Growing}

Progressive growing is an application of transfer learning. In a progressive growing algorithm, layers are added to the generator and discriminator as training moves forward. As layers are added to the networks, generated images increase in spatial resolution. While all layers remain trainable throughout the training period, progressive growing allows $G$ and $D$ to learn increasingly fine scaled features on increasingly high resolution images. Learning step by step presents a series of simpler tasks to the model. Progressive training is consequently more stable and more efficient than traditional training.

Progressive learning takes advantage of a deep neural networks' ability to learn features from generic to specific, or low to high resolution. At each progressive step the weights learned for all the layers in the last step are transfered to identical layers in the next step. This transfer leaves only one untrained layer at each step. By progressively adding layers the network learns the features at each resolution independently, easing the learning task of each progressive network. We employ this technique to produce masks that mirror the input high resolution in sharpness. Figure 4 shows examples of progressively higher resolution images and masks used to train each layer.

Traditionally, progressive GANs are employed for generative tasks. We, however, seek to apply it to translation, specifically segmentation. By using an encoder-decoder structure in the generator we rely on the encoder to map the high resolution input to a latent vector which is translated by the decoder. Like in traditional progressively growing GANs, the decoder is progressively trained. Because we desire the decoder to decode from a latent vector containing all the information contained in the high spacial resolution of our input, the encoder is not progressively grown. The encoder instead maintains its full structure throughout the progressive training.

The discriminator grows in sequence with the decoder. This trains each successive layer to discriminate specific resolutions.

\section{Experimental Evaluation}\label{exp} 

In this section we compare the results of our progressive GAN model to results from a standard U-Net model and GAN model that is not progressively trained. We compare the results both visually and numerically by taking the per-pixel error of the masks.

\begin{figure*}[t!]
\centering
\includegraphics[width=0.9\textwidth]{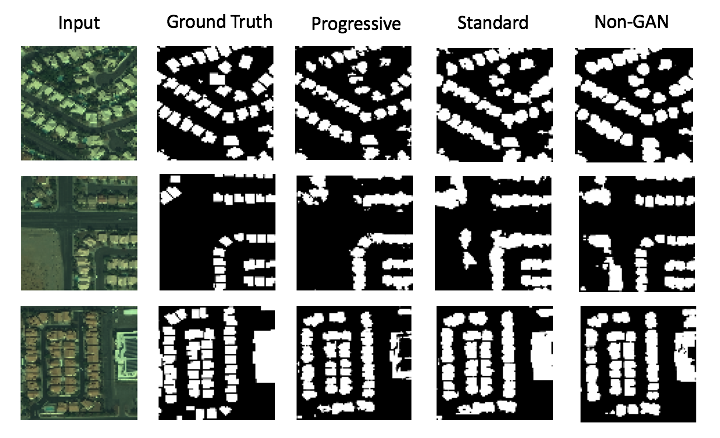}
\caption{Results from the three methods tested along with the input and the ground truth mask.}
\end{figure*}

\subsection{Implementation Details}
For our experiment each model is trained over 32,000 iterations. Training takes places using four NVIDIA Titan X GPUs with a batch size 64 and learning rate 0.0002. Training and testing sets are randomly divided with 70\% of data for training and 30\% for testing. The U-Net model has an encoder-decoder architecture. The encoder is built of 8 hidden layers with 64, 128, 256, 512, 512, 512, 512 and 512 hidden units per layer. The decoder is built of 8 hidden layers that mirror the encoder. This U-Net is embedded as the generator in the GAN. The GAN discriminator is built with the same architecture as the decoder, and grows in conjunction to it. Batch normalization (momentum=0.9 and epsilon=1e-9) and dropout (p=0.5) are employed during training to discourage overfitting.

\subsection{Results}

\begin{table}[h!]
\textbf{Table 1}. Summary of testing and training performance for U-Net, GAN and Progressive GAN.
  \begin{center}
    \begin{tabular}{l|c|c|c}
      \textbf{ } & \textbf{U-Net} & \textbf{GAN} & \textbf{Progressive GAN} \\
      \hline
      Training accuracy & 0.87 & 0.91 & 0.94\\
      Testing accuracy & 0.85 & 0.89 & 0.93 \\
    \end{tabular}
  \end{center}
\end{table}

From Figure 5 we can see that model inferences of rooftop location generally match the size and shape of ground truth. While border regions of rooftops leave room for improvement, we show progressive growing improves the definition of individual buildings compare to its counterparts. In the two non-progressive methods we can see that the building segments tend to blend together more than in progressive growing. Additionally, we can see that the standard GAN and non-GAN approaches suffer from false positives where the progressive growing is able to minimize this. 

\begin{figure}[h!]
\centering
\includegraphics[width=0.5\textwidth]{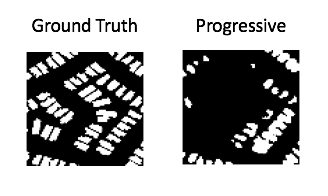}\
\caption{The progressively trained model tends to leave space blank rather than classify possible false positives.}
\end{figure}

One interesting result is the progressive model's ability to limit false positives compared to the other methods. The cause of this is due to the specificity of feature in the later layers. This results in the progressively trained model preferring to not label pixels over commit false positives. Figure 6 presents progressive GAN output demonstrating this phenomenon. As a whole, the progressively trained GAN produces building footprints that snap to the original nicely while also minimizing the amount of false positive pixels compared to the standard methods.

\begin{figure*}
\centering 
\subfigure{\includegraphics[width=59mm]{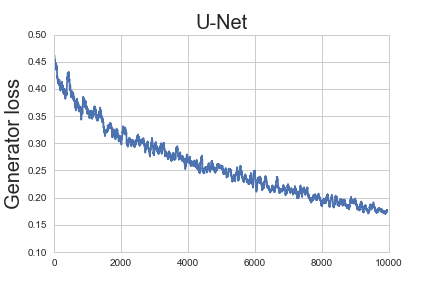}}
\subfigure{\includegraphics[width=59mm]{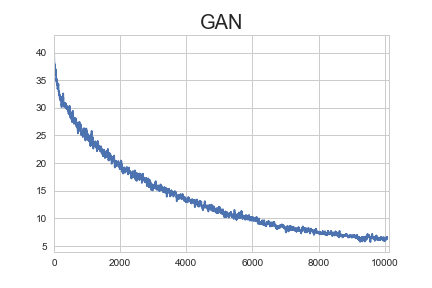}}
\subfigure{\includegraphics[width=59mm]{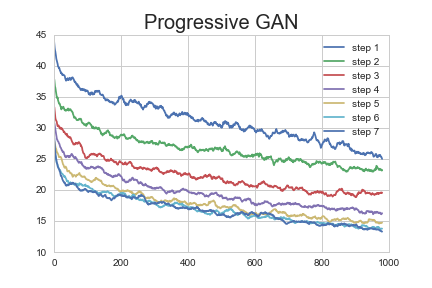}}
\end{figure*}

\begin{figure*}
\centering 
\subfigure{\label{fig:a}\includegraphics[width=59mm]{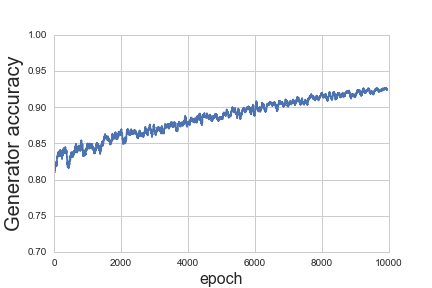}}
\subfigure{\label{fig:b}\includegraphics[width=59mm]{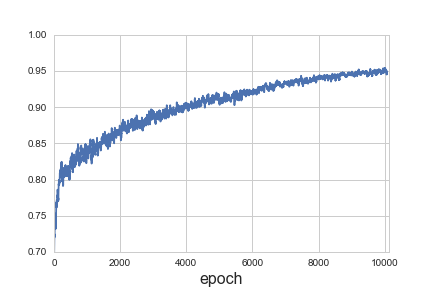}}
\subfigure{\label{fig:b}\includegraphics[width=59mm]{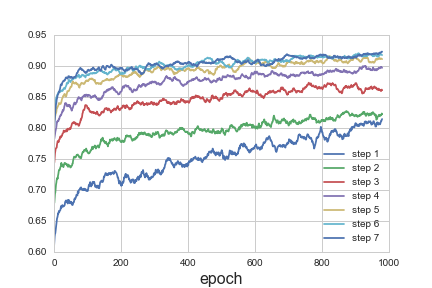}}
\caption{Generator loss and accuracy over training epochs for U-Net, GAN and Progressive GAN. For our proposed model, the progressive GAN, generator accuracy and loss converge to an increasingly better performance with each progressive step until some ceiling is reached at which increasing resolution does not result in learning of finer features.}
\end{figure*}

We present our accuracy scores as the per-pixel error between the ground truth mask and the masks produced by our models. The per-pixel provides a good view of how well the produced masks fit to the high resolution buildings. From Table 1 we can see that the progressively trained GAN outperforms its counterparts in this metric. We also present both the training and testing accuracy of our models to verify that none have over-fit to the dataset. 

In Figure 7 we present graphs for the loss and accuracy of each method during training. We can see that for the progressively trained GAN that each progressive step builds on top of the previous. The decreased loss and quicker convergence at each step shows that there is good transfer of knowledge between the previous and successive steps. Another interesting result is the closeness of the higher resolution layers. This suggests that there exists a certain resolution, in our case 256x256, for which all following resolutions cannot be used to learn increasingly fine features.

\section{Applications in Earth Science}
Multiple scientific and social domains call for high-resolution rooftop products. Production of these products relies on both data collection and accurate methods for semantic segmentation. While some prior efforts have used physics and/or texture-based models, or additional data like elevation data, we focus on an approach relying only on multi-spectral reflectance. Images that are only in the visual spectrum are even more ubiquitous, however a rooftop detection model that uses multi-spectral images is broadly applicable. Multi-spectral satellite images are publicly available from multiple satellites with global coverage and record length stretching back up to over a decade. These images can be used to develop a time series of rooftop data with multiple impactful uses, such as rapid change detection in satellite images in the aftermath of natural disasters.

The results presented in this paper demonstrate an improvement in semantic segmentation performance by GANs using progressive growing. This offers a step toward improvements in high-resolution segmentation. Applications such as 3D city modeling demand high-pixel accuracy rooftop products, and much space for improvement remains. The challenge of training a generalizable rooftop detection model also remains. In addition to variation in appearance of rooftops and non-rooftop area from city to city, buildings vary widely in appearance between highly developed areas and informal settlements like slums. Progressive growing, as an implementation of transfer learning, could offer insights into which learnable features and layers are generally applicable to all cities. With this knowledge, transfer learning could be applied to train models more quickly for rooftop detection in new locations.

\section{Conclusion}\label{con}
This paper presents a novel approach to semantic segmentation that draws upon recently developed machine learning techniques. We evaluate progressive training for semantic segmentation and show improvements upon the performance of prior semantic segmentation approaches. With the increased importance of segmentation in Earth sciences, we present our method for rooftop segmentation on the SpaceNet dataset. 

\section{Acknowledgments}\label{ack}
This project was support by the NASA Earth Exchange (NEX) and NASA CMS grant \#NNH16ZDA001N-CMS.

\bibliographystyle{IEEEtran}  
\bibliography{conference_071817.bib}

\end{document}